\documentclass{article}
\usepackage{arxiv}
\usepackage{setspace}
\usepackage{multirow}
\usepackage{comment}
\usepackage{tabularx}
\usepackage[numbers,sort&compress]{natbib}

\usepackage{graphicx}

\usepackage{algorithm}
\usepackage{algpseudocode}
\usepackage{adjustbox}
\usepackage{hyperref}

\usepackage{mathtools}

\usepackage{adjustbox}

  \usepackage[caption=false,font=normalsize,labelfont=sf,textfont=sf]{subfig}

\usepackage{url}

\title{Bag of Visual Words (BoVW) with Deep Features - Patch Classification \\
Model for Limited Dataset of Breast Tumours}

\author{
 Suvidha Tripathi \\
  Department of Information Technology\\
  Indian Institute of Information Technology Allahabad\\
  Jhalwa, Deoghat, Prayagraj, Uttar Pradesh 211015 \\
  \texttt{suvitri24@gmail.com} \\
   \And
 Satish Kumar Singh \\
  Department of Information Technology\\
  Indian Institute of Information Technology Allahabad\\
  Jhalwa, Deoghat, Prayagraj, Uttar Pradesh 211015 \\
  \texttt{sk.singh@iiita.ac.in} \\
  \And
    Lee Hwee Kuan\\
    1. School of Computing, National University of Singapore,13 Computing Drive, 117417, Singapore\\
    2. Bioinformatics Institute, A*STAR, 30 Biopolis Street, 138671, Singapore,\\
    3. Image and Pervasive Access Lab(IPAL), CNRS UMI 2955, 1 Fusionopolis Way, 138632, Singapore,\\
    4. Singapore Eye Research Institute, 20 College Road, 169856, Singapore\\
    \texttt{leehk@bii.a-star.edu.sg}\\
}

\begin{document}
\maketitle

\begin{abstract}
Currently, the computational complexity limits the training of high resolution gigapixel images using Convolutional Neural Networks. Therefore, such images are divided into patches or tiles. Since, these high resolution patches are encoded with discriminative information therefore;   CNNs are trained on these patches to perform patch-level predictions. It is a most sought-after approach for solving gigapixel classification and segmentation problems. Patch level predictions could also be used to detect new regions in the biopsy samples taken from the same patient at the same site. In other words, once the patient slide has been trained with the model, biopsy samples taken at a later stage to find whether the tumour has spread could be tested using the trained model. However, the problem with patch-level prediction is that pathologist generally annotates at image-level and not at patch level. Therefore, certain patches within the image, since not annotated finely by the experts, may not contain enough class-relevant features. Moreover, standard CNN generate a high dimensional feature set per image, if we train CNN just by patch level, it may pick up features specific to the training set and hence, not generalizable. Through this work, we tried to incorporate patch descriptive capability within the deep framework by using Bag of Visual Words (BoVW) as a kind of regularisation to improve generalizability. BoVW- a traditional handcrafted feature descriptor is known to show good descriptive capability along with feature interpretability. Using this hypothesis, we aim to build a patch based classifier to discriminate between four classes of breast biopsy image patches (normal, benign, \textit{In situ} carcinoma, invasive carcinoma) in respect to better patch discriminative features embedded within the classification pipeline. The aim is to incorporate quality deep features using CNN to describe relevant information in the images while simultaneously discarding irrelevant information using Bag of Visual Words (BoVW). The proposed method passes patches obtained from WSI and microscopy images through pre-trained CNN to extract features. BoVW is used as a feature selector to select most discriminative features among the CNN features.  Finally, the selected feature sets are classified as one of the four classes. The hybrid model provides flexibility in terms of choice of pre-trained models for feature extraction. Moreover, the pipeline is end-to-end since it does not require post processing of patch predictions to select discriminative patches. We compared our observations with state-of-the-art methods like ResNet50, DenseNet169, and InceptionV3 on the BACH-2018 challenge dataset. Our proposed method shows better performance than all the three methods. 
\end{abstract}


\section{Introduction}

Medical image processing has been a challenging domain for the computer vision researchers because of their inherent complex nature and overlapping features/boundaries between classes. A malign tissue mass may get overlooked by an experienced pathologist, whereas a benign mass may get labeled as suspicious because of their highly variable characteristics. For example, in neuroendocrine tumours, it is usually challenging to segregate benign and malignant region because a predominently benign looking neuroendocrine tumour which has the possibility of developing into metastatic cancer can be overlooked by the pathologist due to their weak metastatic morphological signatures \cite{waldum2008classification}. Other prominant challenges like variable staining and acquisition techniques between different dataset images \cite{apple2016sentinel}, inter and intra-observer variations among pathologists \cite{connolly2006changes, cserni2008variations, vestjens2012relevant}, different patterns of cell division, histological and cytological changes, and various other structural changes do not necessarily reflect distinct properties exhibited by the early stages of tumours or by regions which are in proximity of high-grade tumours.
\par
\begin{figure}[htbp]
\centering
\includegraphics[width=5in ,height=3in]{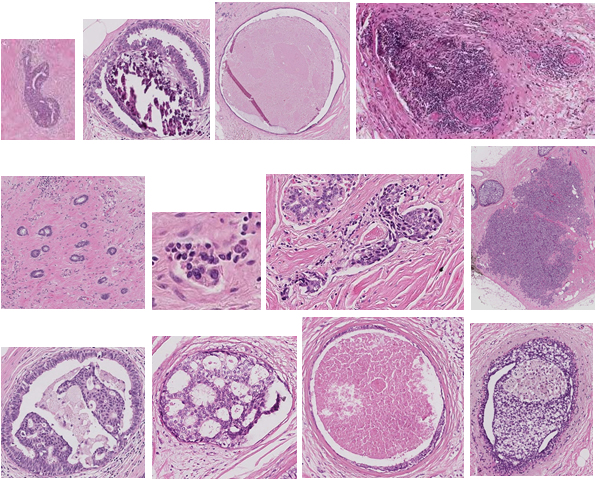}
\caption{Sampled Regions from Whole Slide Images. First row to Third: Benign Tumours, Invasive Carcinoma, \textit{In situ} Carcinoma. Rest of the un-annotated regions were considered as normal.These regions can be seen having different dimensions but represent a single resolution level (level 0) from the WSI pyramid (Fig. \ref{WSIpyramid})}
\label{regions}
\end{figure}

 Further, for WSI, multi-level resolution and hence, high dimensionality adds more challenges like data processing and extraction before we use the multi-layer deep model for further processing. The gigapixel images are necessarily required to be broken into patches for computational analysis through a machine learning model. Most of the classification and segmentation models hence process patches extracted from the high resolution WSIs for diagnosis purposes \cite{spanhol2016breast, gecer2018detection, huang2011time, doyle2006boosting, basavanhally2013multi, bejnordi2016automated, bahlmann2012automated}. Therefore building strong patch-level classifiers is important for developing efficient diagnostic models. However, the patches, since not comprehensively annotated by pathologists at patch-level may exhibit redundant or overlapping features of another class found in the proximity. The \autoref{regions} shows the example regions from the dataset used in this work. We could clearly deduce the highly variable patterns within classes. 
\par To analyse such complex data, recently, multitude of methods have based their CAD models on deep learning frameworks which are prominently Convolutional Neural Networks \cite{
wang2018breast,bandi2018detection,kausar2019hwdcnn, jonnalagedda2018regular, mahbod2018breast, graham2018classification}. Convolutional Neural Networks are better feature extractors than traditional computer vision algorithms and with more data they have proved to scale effectively than classical ML algorithms.  Deep learning methods do not require complex feature engineering and are also adaptable and transferable. However, with all these advantages, they are non-interpretable and therefore, the features generated by deep networks cannot be described in respect to the data. The underlying mechanism to extract features does not allow data specific tuning of features and hence it becomes difficult to analyse the generated features. Hence, directly feeding raw medical images as we do in the case of ImageNet dataset \citep{imagenet_cvpr09} to the deep model does not guarantee efficiency and performance. The same has been observed and proved through our experiments in \cite{tripathi2018histopathological} that state of the art deep learning models alone should not be relied upon for the tasks involving complex medical datasets. 
\par Traditional handcrafted feature engineering methods on the other hand works well with small dataset, they are computationally cheap to process and the most importantly, the features generated are interpretable and easier to understand. We have known for quite a while that medical images are usually analysed by pathologists through their structural changes and perturbations that could only be understood visually. Hence, to realise such structural and textural changes in the computational framework, traditional methods offer advantage over Convolutional Neural Networks since they allow the visual interpretaions to be converted into mathematical simulations more effectively \cite{ozdemir2011resampling, difranco2011ensemble, monaco2010high, krishnan2012hybrid, doyle2008automated, krishnan2012automated, wang2015exploring, caicedo2009histopathology, kather2016multi}.
\par We could see that each domain exhibits its unique challenges and advantages and therefore, the hypothesis that if we could somehow integrate the advantages of both deep learning and handcrafted descriptors, the limitations of patch level analysis could be effectively mapped into an automatic classification network. In this work, the dataset we have used is limited and the patches are not extensively annotated. Combination of these challenges  encouraged the use of BoVW with CNNs. We tested the hypothesis that BoVW could regularise the features extracted by the backbone CNN network in an end-to-end pipeline. In a traditional use of BoVW, local feature descriptors like SURF is used for feature extraction followed by feature selection and vocabulary building. But, it has been constantly proved that CNNs perform better than traditional feature extractors like SURF and SIFT. Therefore, we used this analysis and used CNNs as feature extractors and BoVW as feature selector.

\par The intuition behind using Bag of Features was to select discriminative set of deep features that best describe the image region (formed from the classified patches). Our work is different from other literatures  because we discard irrelevant patch-level features and improve patch classifier within an end-to-end pipeline instead of post-processing patch-level results for selecting discriminative patches like the authors did in \cite{hou2016patch}. Through the proposed work, we focussed on building an efficient patch classification module that could be used within a higher level system for classifying medical images in a larger context of Whole Slide Images. The simple implementation may also help in better clinical workflow integration with sophisticated automated systems.

The contributions of this paper are: 

\begin{enumerate}
\item Our work proves that the combination of deep learning with classical machine learning and feature engineering models produce better observations in case of small, multi-class, and highly complex medical histology datasets. 
\item Multiresolution analysis has been done to extend the viability of the proposed method at different resolution levels of Whole Slide Images. 
\item  Two different datasets, Microscopy images, and WSIs from different patients and acquisition methods have been tested to prove the robustness of our proposed method.
\item The proposed method use pre-trained architecture since the dataset is small. Hence, bottlenecks like huge training time and storage space are eliminated.  
\item Our method is simple and efficient as it does not require heavy preprocessing steps such as stain and color normalization and data augmentation. 
\end{enumerate}
\label{sec1} 
\section{Related Work}
Recently, most research experiments on histopathological image classification has used deep learning-based methods. Most of them have used pre-trained deep models and either fine-tuned them on their dataset or transfer learned the extracted features using traditional machine learning models like Multi-Layer Perceptron, SVMs \cite{fondon2018automatic,nanni2018ensemble}, Decision Trees, Random Forest \cite{lee2018robust}, and other common classifiers. The changing modality or category of images such as microscopy or whole-slide images, used for classification in case of breast cancer assessment, also affects the overall methodology of the problem. For classification of microscopy images in histopathology, there are publically available datasets like BreakHis \cite{spanhol2015dataset}, PatchCamelyon(PCam) \cite{veeling2018rotation, bejnordi2017diagnostic}, BACH \cite{aresta2019bach,bach2018} which provide binary or multi-class labels for classification.  The available literature has tested these datasets with both deep learning-based methods and combination of deep learning and handcrafted features. Most of the time, these microscopy images have high resolution that needs to be broken down into patches for model processing. For instance, the authors in \cite{spanhol2016breast} used BreakHis dataset for classification of malignant and benign tumours using AlexNet based Convolutional Neural Network. They divided their images in the form of patches, and then the classification scores of each patch obtained through CNN were combined to calculate image Level scores. Often, dividing images into patches is the only way to process high-resolution images without losing their magnification and structural features of fine details in the image, such as cells and nuclei. 
\par However, at the same time, the overall features of the boundary and extent of the tumour are also lost.  Image resizing, interpolation, and other types of transformations are some other pre-processing methods to process microscopy images.  Juanying et al. \cite{xie2019deep} used BreakHis dataset in their model by using transformations of original images to balance their dataset and resized them to match the input size of InceptionV3 and Inception\_ResNet\_V2 \cite{szegedy2017inception} models for analyzing breast cancer images. They used transfer learning technique to extract features of these fine-tuned models  (pre-trained on ImageNet) and then classified the features using clustering analyses. Authors in \cite{han2017breast} classified breast cancer images having multi-class labels and instead of binary classification models, provide a more qualitative assessment of breast tissues. They used the novel end to end structured convolutional neural network framework, which uses distance constraint feature space by using prior knowledge of inter and intra-class labels. This constraint improved the learning capabilities of their proposed deep learning model.  They analyzed their results at different magnification levels and achieved the highest accuracy of 93.7\% at image level on their augmented dataset and 94.7\% accuracy at the patient level. Many other studies like \cite{araujo2017classification, sirinukunwattana2016locality} have also used custom CNN architectures using the custom number of layers and hyperparameters such that they perform better on their histopathological datasets. 
\par There is another section of studies which have used Whole Slide Images as their primary dataset. These studies have extracted affected tissue regions from high-resolution WSIs. instead of using microscopy images. Some have first use the detection algorithms to identify regions of interests such as authors in \cite{gecer2018detection} used a saliency detector consisting of a pipeline comprising four Fully Convolutional Networks (FCNs) to detect and extract ROIs in WSIs. They have proposed an efficient detector that analyses multi-scale or multi-resolution images of WSIs. to give the relevant regions for further classification. Such studies are useful to discard white regions in tissue samples along with normal regions with fewer nuclei density which forms most of the regions of WSI. This process, in return, reduces the number of diagnostically irrelevant patches while simultaneously reducing computational cost. Similar studies in \cite{huang2011time, doyle2006boosting, basavanhally2013multi, bejnordi2016automated, bahlmann2012automated} have proposed different methods to detect relevant regions for suspected cancerous regions in different types of tissues. Their methodologies try to quantify the pathologists' view of interpreting slides hierarchically from lowest resolution to highest resolution.   After the detection step, a further process of classifying selected tissue sections as malignant or benign has been the same as classifying microscopy images where the large images are broken down into fixed-sized patches. These patches are then classified using various deep learning networks to the final results are combined to give both patch level and slide level classification. Commonly used deep learning models for this purpose are ResNet and its variants, InceptionV2, InceptionV3, a combination of ResNet and Inception such as Inception\_ResNet\_v2 model used by authors of \cite{xie2019deep}. \par Recently, The BACH 2018 challenge concluded by Guilherme Aresta et al. \cite{aresta2019bach} has listed the type of deep learning models used by participants across the world in their paper \cite{aresta2019bach}. The top 3 contenders  used ResNet-34,50,101, DenseNet161 \cite{huang2017densely}, Inception-ResNet-v2, and InceptionV3. Other architectures most often used by participants of BACH challenge were XCeption \cite{chollet2017xception}, GoogleNet, VGG16-VGG19 \cite{simonyan2014very}. This paper \cite{aresta2019bach} showed that with the proper choice of hyper-parameters and fine-tuning the patch classification could be achieved with high accuracy using pre-trained deep models. The maximum accuracy achieved by the top participant was 87\% with ResNet101 and DenseNet161 model.  It was important to note that most of these models have been pre-trained on ImageNet. However, all these methods lacked feature interpretability which is very crucial for identifying the features responsible for discriminating normal regions from cancer. Without interpretability, the research scope for improving these methods gets limited to developing either custom models with different layers and hyperparameters or using these pre-trained models with some pre and post-processing techniques. We know that handcrafted features provide us with the tool of better feature understanding by extracting domain-specific local features. These local features, in case of WSIs or microscopy images, act as a discriminating factor among classes. For instance, a research experiment by  \cite{zhang2018classification} combined handcrafted feature descriptors such as BoF (Bag of Features) and LBP (Local Binary Pattern) with deep models such as VGG-f, VGG19, and Caffe-ref to construct an ensemble model to classify medical images in biomedical literature. Similarly, \cite{wang2014mitosis} detected mitosis in breast cancer pathology by combining morphology, color and texture features with light custom CNN model. Many other studies like \cite{ACM} conducted experiments holding on to similar hypothesis of combining handcrafted and deep features to prove that such methods help to integrate the interpretability of handcrafted features and generalizing capability of deep features and hence produce better results. Such hybrid models have one common limitation that their pipeline is not end-to-end and the discriminative factor imbibed by these methods through handcrafted feature algorithms is concatenated with CNN features. They all required deep framework training for feature extraction and then subsequently combined handcrafted features for classification. Furthermore, most of these pieces of literatures did not evaluate their proposed models with benchmark methods and hence did not satisfactorily justify their choice of methodology. 
\label{sec2}
\section{Methodology}
\label{sec3}
\subsection{\textbf{Overview}}
\begin{figure*}[htbp]
\centering
\includegraphics[width=\textwidth, height=7in]{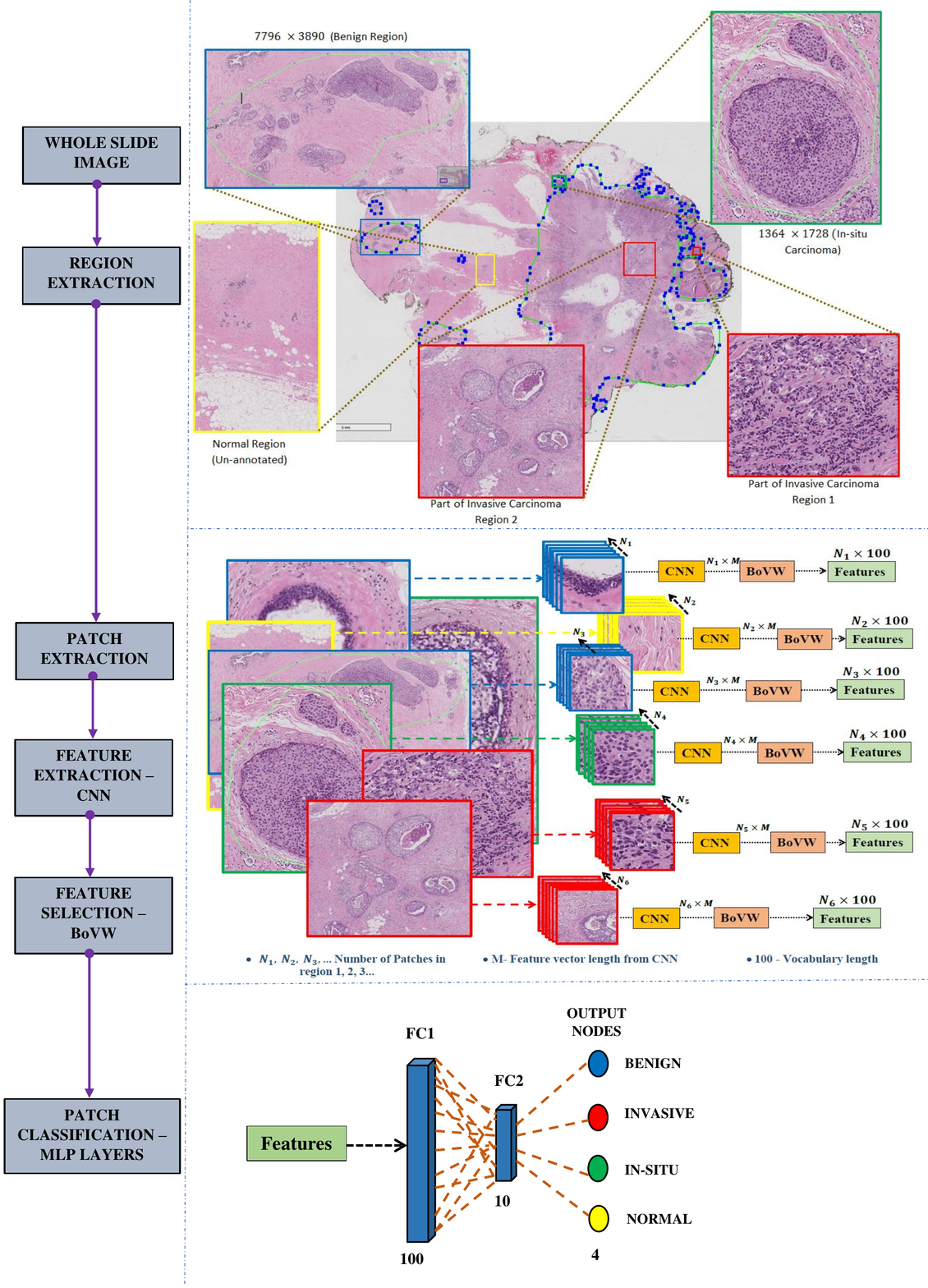}
\caption{The figure illustrates the complete methodology, including the database preparation from Whole slide images. The left-hand side shows the complete flow of the steps in the direction of the arrow, whereas the right-hand side shows the pictorial representation of the steps in detail. The Process starts with extracting rectangular regions around the annotated tumours in WSI. It is followed by patch extraction from the annotated regions, feature extraction through CNN and feature selection by BoVW. The final features obtained from each set is then forwarded for MLP training and classification, which comprise the last step of the proposed method. In the figure, $N_1$, $N_2$, $N_3$, $\ldots$ are the number of patches extracted from each region and $M$ is the feature length obtained from pre-trained CNN. }
\label{patchClass}
\end{figure*}

Patch level analysis is the most followed approach to process high resolution histopathology images in order to avoid losing structural details present in high-resolution images. Pathologists look for cellular-level visual features to determine cancer subtypes and the clear features could only be identified at patch-level scale. Since image-level processing in case of large dimensional microscopy and WSIs would require heavy resizing therefore, our purpose is to build a high performing patch-level classifier instead of image-level classifier. We propose an end-to-end classification pipeline to classify four types of histology patches, namely Normal, Benign, \textit{In situ} Carcinoma, and Invasive Carcinoma present in WSIs and microscopy dataset. Figure \ref{patchClass} gives an overview of the complete methodology. The WSI dataset comprise ten annotated Whole Slide Images. Each WSI is annotated by experts for three classes as mentioned and all the un-annotated parts of the WSI are considered as normal. The pipeline starts with extracting annotated tumour regions from each WSI in the dataset. The extracted regions were then divided into non-overlapping patches. The patches obtained from one region formed a set. These sets contain patches from the single region only. These sets move forward for the feature extraction process.   We have known that state of the art deep learning algorithms like AlexNet \cite{krizhevsky2012imagenet}, VGG16 and VGG19 \cite{simonyan2014very}, GoogleNet \cite{szegedy2015going}, extract  deep features that are capable of producing comparable classification results on ImageNet data \cite{imagenet_cvpr09}. Utilizing this information, we used a pre-trained GoogleNet (ImageNet) model to extract deep features. We advanced further with feature post-processing using Bag of visual words. Processed features were then classified using Artificial Neural Networks (ANN). This classification model was developed, keeping in mind the overlapping features between classes. Clustering similar key-points in a region and discarding weak features in the images resulted in a refined feature set with less variance and hence lower fluctuations. Proposed classification model treated histology images as a set of non-overlapping patches that collectively comprised the whole region. Lastly, we trained  MLP on the distribution of the quantized features in an image. Three broad steps of the pipeline are (i) Pre-processing, (ii) Feature extraction and selection, (iii) Classification. 
\subsection{\textbf{Pre-processing steps}}
\subsubsection{Region Extraction}
\label{sec3.1}
\begin{figure}[htbp]
\centering
\includegraphics[width=\textwidth ,height=3in]{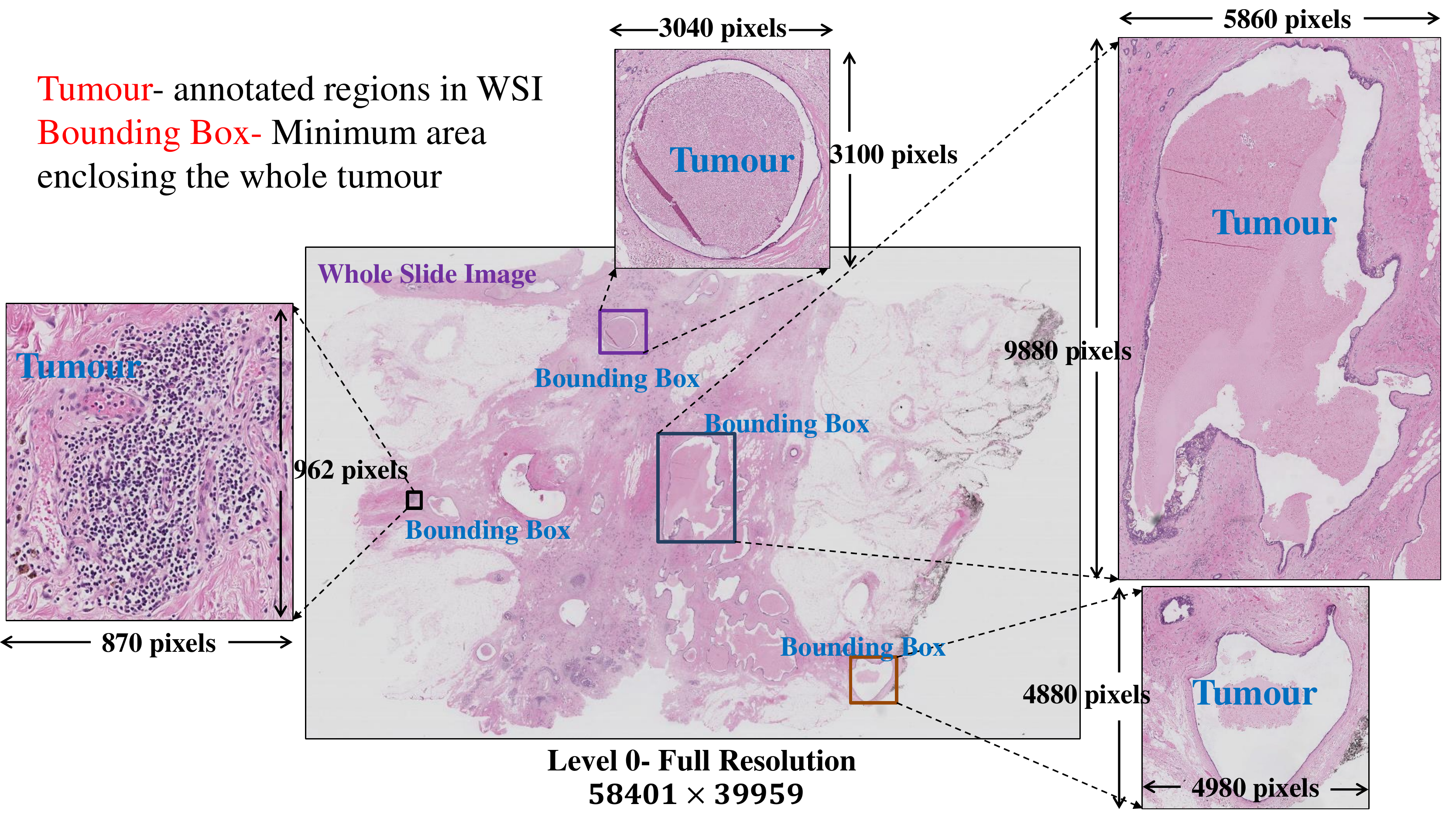}
\caption{Illustration of region extraction process from a Whole Slide Image. The annotations by the pathologists are used to automatically extract the enclosed bounding box area of the annotated tumours in the Whole Slide Image.}
\label{dataset}
\end{figure}

The 10 WSIs comprised 57 Benign tumours, 109 Invasive carcinoma tumours, and 60 \textit{In situ} carcinomas. From each WSI, the ground truth coordinates of annotated polygons were used to extract the bounding box information of the enclosed tumour. The bounding box around the annotated polygon contains the tumour region we want to classify (Figure \ref{dataset}). At the highest resolution level where the dimensions of a WSI may vary between thirty thousand pixels to hundred thousand pixels along each dimension, few annotated tumour regions exceeded the space limitations of the .png image file format. Such regions could not be extracted from the WSis, since they could not be converted into the image format feasible for storing in the disk. After analysing all the Whole Slide Images, seven such invasive carcinoma regions were found to be above the space limitation, hence total 102 invasive regions out of total 109 were finally extracted for further processing.  All extracted regions from 10 WSIs were of arbitrary dimensions ranging from 20570 to 195 pixels across width and 17290 to 226 pixels across height. In order to make the process more robust and rotation invariant, we rotated all regions such that the vertical axis is the first principle component and then cropped bounding box around the rotated region. The steps are as follows:
\begin{enumerate}
\item Using the mask of the annotated regions we calculated the major axis length and the orientation angle of the major axis from the X-axis.
\item We rotated all the regions around the centroid of the major axis by the angle = 90-major axis angle.
\item Once the region mask and the ground-truth are rotated, we calculated the bounding box coordinates of the mask. 
\item Bounding box dimensions calculated around the annotated regions were extended to make the size of the bounding box equal to the closest multiple of 256. 
\item The rotated regions were then cropped around the obtained bounding box coordinates. 
\end{enumerate}

\subsubsection{Patch Extraction and Set Formation}
\begin{figure}[htbp]
\centering
\includegraphics[width=0.8\textwidth ,height=3in]{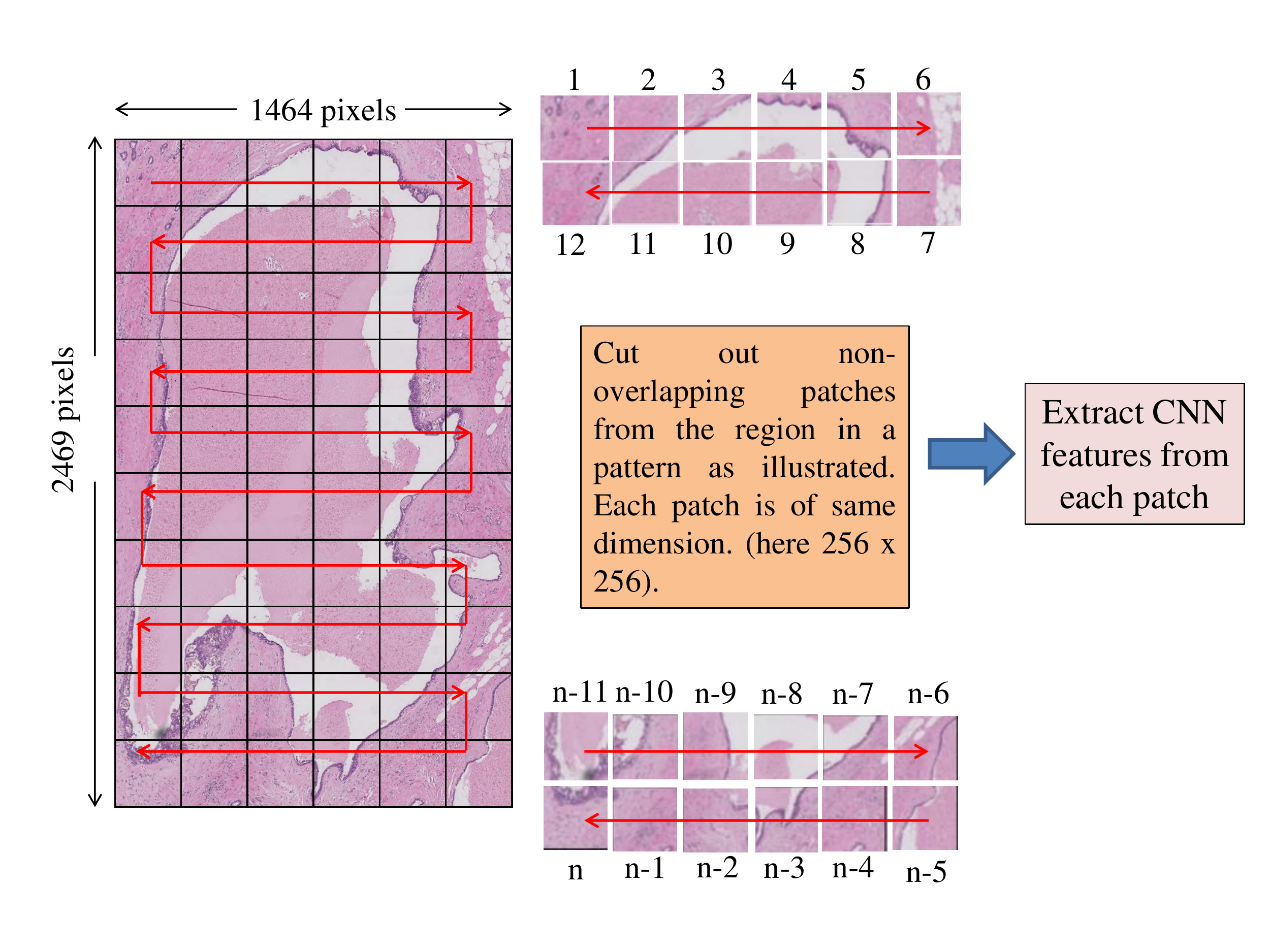}
\caption{Representation of patch extraction procedure for tumour classification. The extracted Tumour regions from each WSI are further divided into patches after the pre-processing steps followed in \autoref{sec3.1}. The patches are extracted following the pattern explained in the Figure.}
\label{patchExtractLSTM}
\end{figure}

The variations in the size of each region made them impossible to feed into a deep learning algorithm which requires every image to be of the same size. Due to acute effects in image resolution and features, we discarded the choice of resizing such vast range images. In the case of medical images resizing should not be considered as an option if the difference in sizes between images is vast. Hence, we divided the tumour regions into non-overlapping patches. We used a non-overlapping sliding window protocol in which the window of size $256 \times 256$  was slid over the image dimensions with stride 256 to extract the patch from the image. The patches from the rotated masks regions of corresponding RGB images were also extracted in the same manner. The mask patches helped us identify regions that were within the bounding box area but not enclosed within the mask, hence we discarded those mask patches and their corresponding RGB patches that have more than 20\% of the area as background.
The pattern to move the sliding window was formulated so that we could obtain the patches in some ordered way. By extracting patches in as ordered way as possible in histopathological images, the drawback of high dimension could be turned into an advantage.  Therefore, using the sliding window, the patches are cut out in a sequence as described in Figure \ref{patchExtractLSTM}. The sequence of scan is from left to right, down, and then right to the left. In the Figure, the direction of red arrows and the numbering of patches show the intended sequence.  left to right $1->2->3->4->5->6$ then down. Then left to right $7->8->9->10->11->12...$ and so on. The whole sequence chain is $1->2->3->4->5->6->7->8->9->10->11->12.....m_n$, where $m$ is the number of patches in the $nth$ region. The patches extracted in this manner form a certain sequence such that when rearranged following the same pattern would give the original image. The method of extraction could be extended in some more efficient manners such that it maintains as much structural architecture as possible with low and even image dimensions across all regions.   All the ordered patches are then stored in a set for further processing in a manner that the patches are read in sequential order. 
Patches from each tumour region will be accumulated in the corresponding set. Let $S$ be the set of patch sets from $L$ tumour regions such that $S=\{N_1, N_2,\ldots, N_L\}$, where $N_n$ is the $nth$ patch set and $n=\{1,2,3,\ldots,L\}$. Each patch set will have different number of patches. So, let $nth$ patch set has $m$ number of patches symbolically denoted by $m_n$. If $p^n_i$ be the patch $i$ in $nth$ patch set where  $\forall p^n \in N_n; i=1,2,\ldots, m_n$. ;
\par Figures \ref{micro} and \ref{wsi} show the example patches extracted from each class. From the Figure, we could view that without an expert's knowledge, one could not distinguish invasive mass from \textit{In situ} carcinoma and the normal patch may look like a benign mass and vice versa. The annotations are usually unreliable and hence, prone to a high rate of misclassification.

\label{sec3.2}

\subsection{\textbf{Feature Extraction and Selection}}
\label{sec3.3}

\label{sec3.3.1}

\subsubsection{CNN feature extraction}
After forming image sets from each labeled region, image datastore of individual sets proceeded for deep feature extraction, one at a time. We used pre-trained GoogleNet architecture (trained on ImageNet) for feature extraction. The method extracted features one set at a time. Since patches from one set belong to one region and one label, CNN could capture the better correlation between features. Also, with this process, we could feed the whole tumour region into CNN, in its full resolution, without the need for resizing. 
\label{sec3.3.2}
\subsubsection{BoVW Feature Selection}
\begin{figure*}[htbp]
\centering
\includegraphics[width=\textwidth ,height=2.5in]{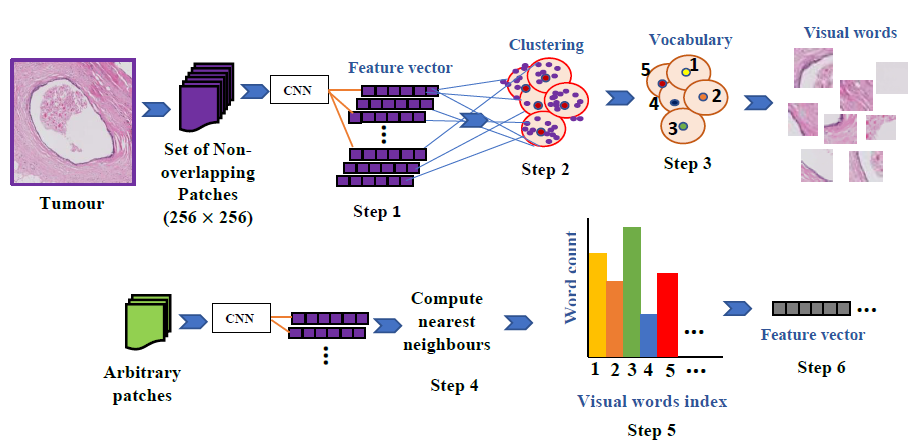}
\caption{BoVW feature selection pipeline: \textbf{Step \#1: Feature Extraction:} Input- Image patch, output- CNN Feature vector. \textbf{Step \#2 \& 3: Vocabulary Construction:} Feature vectors are clustered using the K-means clustering algorithm. The resulting cluster centroids are our reference points to link similar features around the cluster centers. Centroids act as our codebook of visual words.
\textbf{Step \#4 \& 5: Vector quantization:} Extract features from test images, compute nearest neighbors of feature vectors with codebook centroids using Euclidean Distance.  Build a histogram of length k (number of centroids) and plot the features corresponding to their nearest centroid. The centroid having highest frequency would tell that the image relates to this cluster center the most. 
\textbf{Step \#6: Feature vector calculation:}  from the histogram, we calculated the feature vector per image patch with K cluster centers on the x-axis and frequency of each center on the y-axis.}
\label{bovw}
\end{figure*}

Bag of visual words (BoVW) is a commonly used image classification method. The adapted concept is from information retrieval and NLP’s bag of words (BoW) \cite{harris1954distributional}. In the algorithm, the record of the number of times a word appears in a document is maintained. Then, the frequency of every word is calculated to list the keywords of the document. Then, we draw a histogram plot with the word against its frequency. The process then transforms a document into a bag of words (BoW). Following the same concept, in case of images, the algorithm is called a “bag of visual words.” However, instead of words, we use image features as the “words.” Several local patches abstract a representation of an image as a set of features. Features consist of keypoints and descriptors. 
In our work,  CNN feature extraction replaces the feature detection step in Bag of Visual Words algorithm. Traditionally, the feature detection step is applied to detect key features in the image, and a patch around that key feature is extracted for further processing. Now, due to patch-based modeling in our work, the image has been already converted into a set of patches. These set of patches from the image were then directly represented as a set of feature vectors after passing them through CNN. 
We used CNN as our custom feature extractor since CNN is a more generalized non-linear feature extraction framework; therefore differences in the images can be represented more efficiently through CNN than traditionally used keypoints detector descriptors like SIFT and SURF. 
Figure \ref{bovw} shows the visual pipeline of feature extraction and creating a vocabulary of 100 visual features from a set of images. 
\\Features per set are then trimmed to discard weak features The BoVW algorithm selects 80\% of the strongest features for further processing. The BoVW calculates the strongest features from the variance metric calculated from the original feature set. The trimmed feature set is then divided into hundred clusters using k means algorithm \cite{forgy1965cluster}. Selecting differentiating, important, and well-separated key points within a set with the help of BoVW algorithm refined the feature space of a region. BoVW created a visual vocabulary or codebook for the same set features. Similar features can provide an approximate estimate as to what the region is, just as a summary of an article express what the lengthy article says in a few words. BoVW returns a feature bag or codewords, which is a codebook of similar features represented by their cluster centers. We have prepared our codebook of k=100 clusters. Thus each patch in an image is mapped to a certain codeword in the codebook following the clustering process. Next step is to encode the codewords into a histogram representation so that we obtain a feature vector of dimension $D$ from $nth$ set, where
\begin{equation}
\begin{aligned}
D &=N_{n} \times {k} \ni N_{n}=card(\sum^m_i p^n_i),  \\
  & {k}=100
\end{aligned} 
\end{equation}
Histogram defines each image in terms of the generated codebook. The encoding process locates the cluster centroid nearest to the location of the current feature. The feature histogram represents the number of features in each cluster. For the cluster with the highest number of features, we can deduce that the region is represented by that cluster the most in comparison to other clusters. The region is now converted into a selected number of a most representative set of features. After all the steps, the most important part of the feature space formation is completed, which is then followed by training the machine to assign labels to images using Artificial Neural Network. 
\label{sec3.2.3}
\subsubsection{Feature Concatenation}
Selected features from each set $P^n$ are then concatenated with other set features by appending rows  of features such that the total feature set dimension formed after concatenation is represented as:
\begin{equation}
\begin{aligned}
T &=N_{total} \times {k} \ni N_{total}=card(\sum^L_n \sum^m_i p^n_i),  \\
  & {k}=100
\end{aligned} 
\end{equation} 
\label{sec3.3.4}
\subsection{\textbf{Patch Classification}}
Let us perceive our encoded feature vector as the frequency of samples along the x-axis of a histogram and number of significant features against each sample along the y-axis of the histogram. So, in a nutshell, 100 features completely describe each sample. We then trained MLP as our multiclass classifier because of its better performance in such kind of images. Evaluation of the performance of MLP in classifying histopathological features is carried out in \cite{ACM}. Also, the choice of shallow network is feasible for the cases the training dataset is small, as in our case.  
\label{sec3.3.5}

\section{Setup and Results}
\label{sec4}
\subsection{\textbf{ICIAR 2018 BACH Dataset}}
The BACH (\textbf{B}re\textbf{A}st \textbf{C}ancer \textbf{H}istology) dataset was divided in two datasets to solve different problems on each set. The first dataset \textbf{DS1} contained 400 histology microscopy images, 100 from each of the following classes: 1) Normal, 2) Benign, 3) \textit{In situ} Carcinoma, and 4) Invasive Carcinoma. Each of the histology images was $2048 \times 1536$ in dimensions. We did not use any pre-processing steps on DS1. The second dataset \textbf{DS2} contained 10 annotated Whole Slide Breast histology Images containing the entire sampled tissue. These WSIs are high resolution images digitized in \textit{.svs} format and variable size with maximum width and height of 62952, 44889 pixels, respectively. With each WSI file and \textit{.xml} file was released which contained the pixel coordinates of polygons that enclose each labeled region. Every WSI contained more than one pathological regions with different labels. The regions have the same four labels as the histology microscopy slides. The complete dataset details can be found in their paper \cite{aresta2019bach} and the challenge website \cite{bach2018}.
\label{sec4.1} 
\subsection{\textbf{Dataset Preparation}}
\begin{figure}[htbp]
\centering
\includegraphics[width=3in ,height=2in]{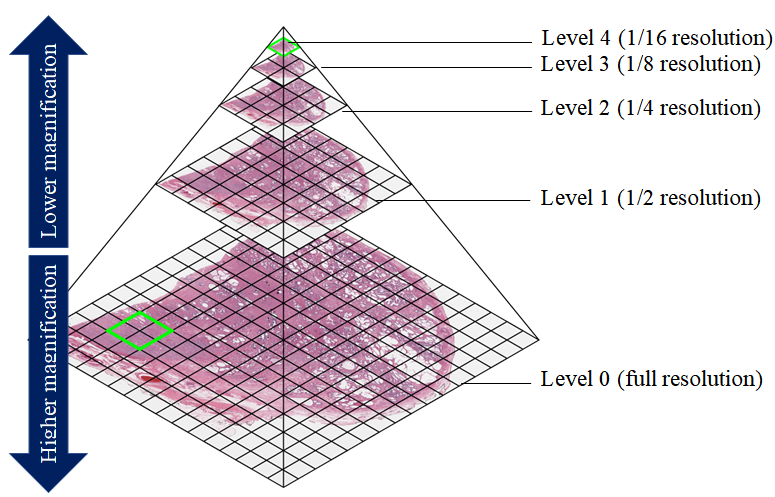}
\caption{WSI pyramid structure. WSI is a multi-resolution file as illustrated in the figure.}
\label{WSIpyramid}
\end{figure}

\begin{figure}[htbp]
\centering
\includegraphics[width=5in ,height=3in]{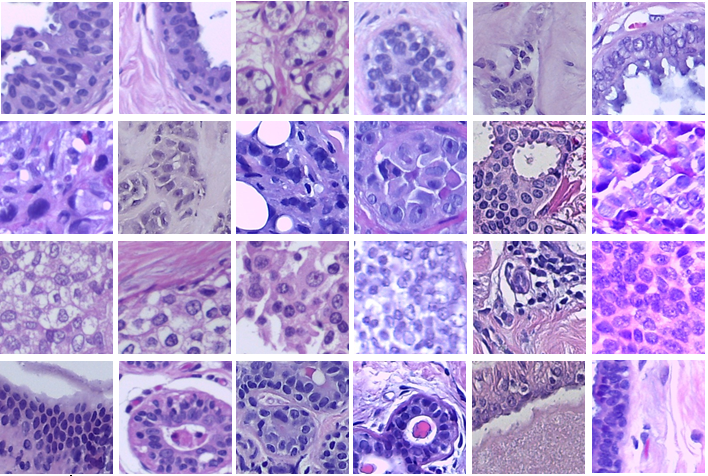}
\caption{Example sub-images of different classes of Microscopy patches of Dataset \textbf{DS1} starting from the first row to fourth:  Benign, Invasive Carcinoma, \textit{In situ} Carcinoma, and Normal}
\label{micro}
\end{figure}

\begin{figure}[htbp]
\centering
\includegraphics[width=5in ,height=3in]{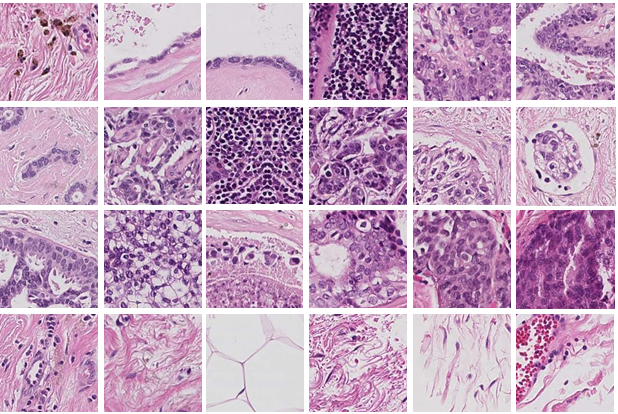}
\caption{Example sub-images of different classes of WSI patches of Dataset \textbf{DS2} starting from the first row to fourth:  Benign, Invasive Carcinoma, \textit{In situ} Carcinoma, and Normal}
\label{wsi}
\end{figure}

For Dataset \textbf{DS1}, the $2048 \times 1536$ microscopy images were divided into a grid of $ 8 \times 6$ dimensions. We call it a grid of patches with each patch of $256 \times 256$ dimensions and total 48 patches were acquired from each histology microscopy image.  Figure \ref{micro} shows the data samples from Dataset \textbf{DS1} segregated by classes. For the second part \textbf{DS2}, we had only 10 annotated WSIs to perform training and testing. From the 10 annotated images, we have  extracted the bounding box region around the perimeter of labeled regions using annotated coordinates information provided in the \textit{.xml} file. Each WSI is digitized at different resolution levels (Figure \ref{WSIpyramid}). We have extracted data from 3 levels. Level 0 is the highest resolution level having a maximum number of pixels and find its place at the lowest pyramid level, as seen in Figure \ref{WSIpyramid}. Level 2 is one-by-fourth the size of level 0, and similarly, level 4 is one-by-fourth of level 2. We henceforth referred these three resolution levels as scale 0, scale 1, and scale 2. From scale 0, we extracted benign, \textit{In situ} carcinoma and invasive carcinoma regions as annotated by pathologists, while normal regions are extracted randomly from the unlabelled tissue section. The total number of samples acquired from scale 0 were 23,851. We followed the same procedure for scale 1 and scale 2 and the number of samples acquired were 8,214 and 1,389, respectively.  The illustration for the annotated region extraction from WSI images in dataset \textbf{DS2} is shown in Figure \ref{dataset}. Figure \ref{wsi} shows the sample data points extracted from 10 WSIs stratified by each class. 

\label{sec4.2}

\subsection{\textbf{Data Usage}}
The number of samples acquired from Dataset \textbf{DS1} containing microscopy histology images was 19,200, whereas 23,851 samples from Dataset \textbf{DS2} of WSIs (Level 0).  DS1 has 4800 instances of each class whereas, the class distribution for DS2 is 7,738 for benign (class 1), 6,444 for invasive carcinoma (class 2), 2,752 for \textit{In situ} carcinoma (class 3), and 6,917 for normal regions (class 4). We did 5 fold cross-validation to test that our results are consistent across the data division.  

\textit{5 fold cross-validation}: Random selection of data samples was made using 5 fold cross-validation method for dividing into training and testing set. The data samples in this method are chosen randomly but with roughly equal proportion of classes in each set. We created five disjoint sets or folds from 19200 samples in Dataset \textbf{DS1} and 23,851 samples in Dataset \textbf{DS2}. We have evaluated the performance of our proposed model using standard metrics such as Precision, Recall, F1-Score and Multiclass AUC \cite{hand2001simple}.  


All the images extracted from microscopy and WSIs had dimensions $256 \times 256$, and they were used in their raw form without any color normalization and adjustment. 
\label{sec4.3}
\subsection{\textbf{Experiments}}
\begin{table*}[htbp]
\renewcommand{\arraystretch}{1.3}
\caption{Comparative Performance parameters for patch wise classification stratified by Dataset. DS1- Microscopy dataset, DS2- WSI dataset}
\label{table1}
\centering
\begin{adjustbox}{width=\textwidth}
\begin{tabular}{|c|c|c|c|c|c|}
\hline
\textbf{Method}&\textbf{Dataset}&\textbf{Precision (\%)}&\textbf{Recall(\%)}&\textbf{F1-Score(\%)}&\textbf{Multiclass AUC(\%)} \\
\hline
\multirow{2}{*}{\textbf{ResNet50 \cite{he2016deep}}}&\textbf{DS1}&73.35 $\pm$ 4.62 & 73.35 $\pm$ 4.62&73.35 $\pm$ 4.62&91.63 $\pm$ 2.45\\
\cline{2-6}
&\textbf{DS2}&70.97 $\pm$ 4.34&70.97 $\pm$ 4.34&70.97 $\pm$ 4.34&90.81 $\pm$ 2.02\\
\hline
\multirow{2}{*}{\textbf{InceptionV3 \cite{szegedy2015rethinking}}}&\textbf{DS1}&79.58 $\pm$ 7.75&79.58 $\pm$ 7.75&79.58 $\pm$ 7.75&94.20 $\pm$ 3.16\\
\cline{2-6}
&\textbf{DS2}&85.60 $\pm$ 6.15&85.60 $\pm$ 6.15&85.60 $\pm$ 6.15&96.89 $\pm$ 1.86\\
\hline
\multirow{2}{*}{\textbf{DenseNet169 \cite{huang2017densely}}}&\textbf{DS1}&80.59 $\pm$ 9.66&80.59 $\pm$ 9.66&80.59 $\pm$ 9.66&94.62 $\pm$ 3.96\\
\cline{2-6}
&\textbf{DS2}&94.37 $\pm$ 1.24&94.37 $\pm$ 1.24&94.37 $\pm$ 1.24&99.36 $\pm$ 0.23\\
\hline
\multirow{1}{*}{\textbf{PBC \citep{roy2019patch}}}&\textbf{DS1}&77.44 $\pm$ *&77.40 $\pm$ *&76.94 $\pm$ *&*\\
\hline
\multirow{2}{*}{\textbf{CNN-BoVW (Proposed)}}&\textbf{DS1}&93.88 $\pm$ 3.03&93.87 $\pm$ 3.03&93.87 $\pm$ 3.03&99.29 $\pm$ 0.64\\
\cline{2-6}
&\textbf{DS2}&98.08 $\pm$ 0.28&97.86 $\pm$ 0.27&97.96 $\pm$ 0.26&99.81 $\pm$ 0.06\\
\hline
\end{tabular}
\end{adjustbox}

* data not reported

\end{table*}
The CNN-BoVW model for patch classification was set up to take tumour patches as input and classify them into four classes: Benign, \textit{In situ} carcinoma, Invasive carcinoma, and Normal. 
\par \textit{\textbf{Experiment 1:}} First, the patches from dataset \textbf{DS1} and dataset \textbf{DS2} were classified separately. The test phase comprised testing sampled patch test set against the trained model. The analysis of the classification performance was done based on Precision, Recall, F1-score, and multiclass-AUC (Area Under the curve). Table \ref{table1} shows the performance metrics with respect to both Datasets \textbf{DS1} and \textbf{DS2}. We calculated the average of the five folds using micro-average method. This method is preferred if there is a class imbalance. For benchmarking our model, we compared the performance of patch classification with recent state of the art deep learning methods- ResNet50 \cite{he2016deep}, DenseNet169 \cite{huang2017densely}, and InceptionV3 \cite{szegedy2015rethinking}.  \\
From the Table \ref{table1}, we could observe that the proposed method has performed better than all the contemporary state of the art deep learning classification networks. These models were pre-trained on ImageNet and later fine-tuned on our datasets. The models were trained with Stochastic Gradient Descent Optimizer with constant learning rate of $10^{-3}$. Each benchmark model was trained for 50 epochs for both datasets. The decision to use pre-trained networks was to eliminate the drawback of fewer data samples. Since these networks need a huge amount of data to generalize, the lack of enough data points in our case, prompted us to utilize the fine-tuning method. Most of the current research in medical domain justifies the use of pre-trained networks on ImageNet data such as in the survey published by ICIAR2018 competition organizers \citep{aresta2019bach} where the top contenders have published their methods using pre-trained models trained on ImageNet. Microscopy images in Dataset \textbf{DS1} (refer Fig. \ref{micro}) are focussed tumour regions containing the class nuclei structures which cover the complete image dimensions.  However, DS1 does not contain tumour structural information. Whereas, in the case of DS2 (Fig. \ref{wsi}), the care had been taken to acquire as much focussed tumour area as possible, but without losing structural information present in boundary patches. Hence, DS2 contains the tumour structure information along with the nuclear details. Therefore, the results for Dataset \textbf{DS2} are much better than Dataset \textbf{DS1} across all the models mentioned in Table \ref{table1}. Recently published study on the dataset \textbf{DS1} by Kaushiki et al. \cite{roy2019patch} proposed Patch Based Classifier (PBC). Their method extracts the patches of size $512 \times 512$ using non-overlapping window. They used stain normalization and patch augmentation as post-processing steps before passing the patches for classification in their proposed custom CNN network. They achieved the average classification accuracy of 77.4\% in four classes. They did not use Dataset \textbf{DS2} for their method testing. After comparison with the models described in Table \ref{table1},  our proposed model have performed better for both datasets since it selects the strong key features among the region patches and discard features that have a high variance from the majority of the features. Whereas, all the other models from which we have shown the comparison do not incorporate discriminative ability within their network. We have used regularization parameter of 0.2 for all datasets to bring uniformity in the training process for training selected features from BoVW output with a neural network classifier. 

\par \textit{\textbf{Experiment 2:}} In a separate experiment, the patches from the two datasets \textbf{DS1} and \textbf{DS2} were merged into one (Dataset \textbf{DS1 + DS2}) to increase the number of samples. The total number of samples is now 43051 divided into four classes - Benign (12538), Invasive Carcinoma (11244), \textit{In situ} carcinoma (7552),  and Normal (11717) patches. Table \ref{table2} shows the comparative performance of deep learning models with our proposed model. From the table, we observe that DenseNet169 and InceptionV3 have given a better performance than the proposed method in this case. The reason may be due to the presence of high heterogeneity among datasets and hence intra-class instances from the two datasets might have caused BoVW to select different vocabulary words with greater intra-class distance. This reduction in performance may also be attributed to the high color variations among the two datasets. The staining and scanning process of two datasets do not conform to one standard, and hence it increases the differences in tissue samples. These differences occur within a class which affects the generalization of the model.  

\begin{figure*}
\centering
\includegraphics[width=\textwidth ,height=3in]{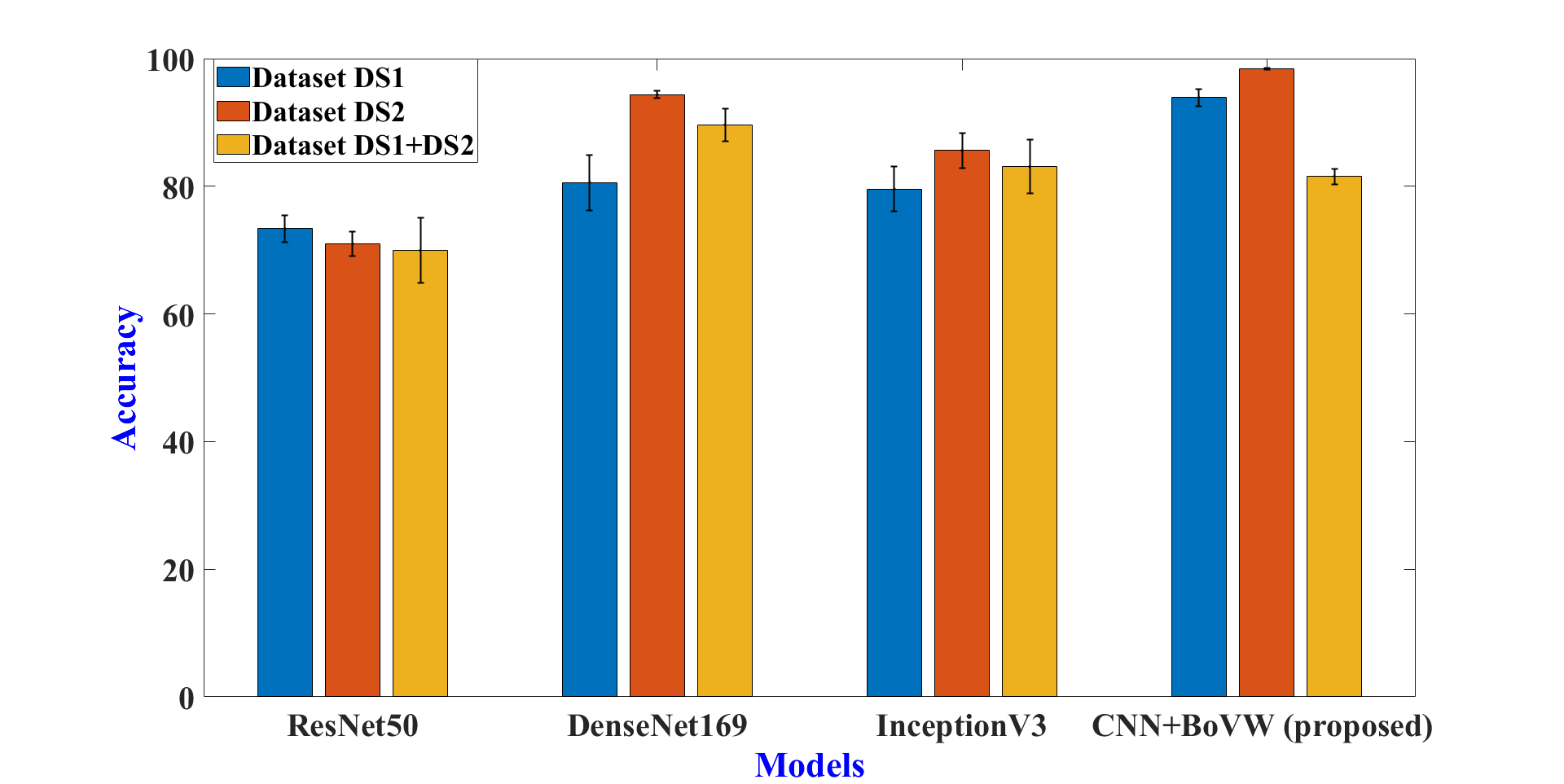}
\caption{Comparative patch wise classification Accuracy obtained with respect to each dataset. DS1- Microscopy dataset, DS2- WSI dataset. The accuracies are plotted as mean accuracy on y-axis with error bars showing standard error obtained after five fold cross validation. The accuracies are obtained after Experiments 1 and 2 as explained in the text}
\label{subfig:fig1}

\includegraphics[width=\textwidth ,height=3in]{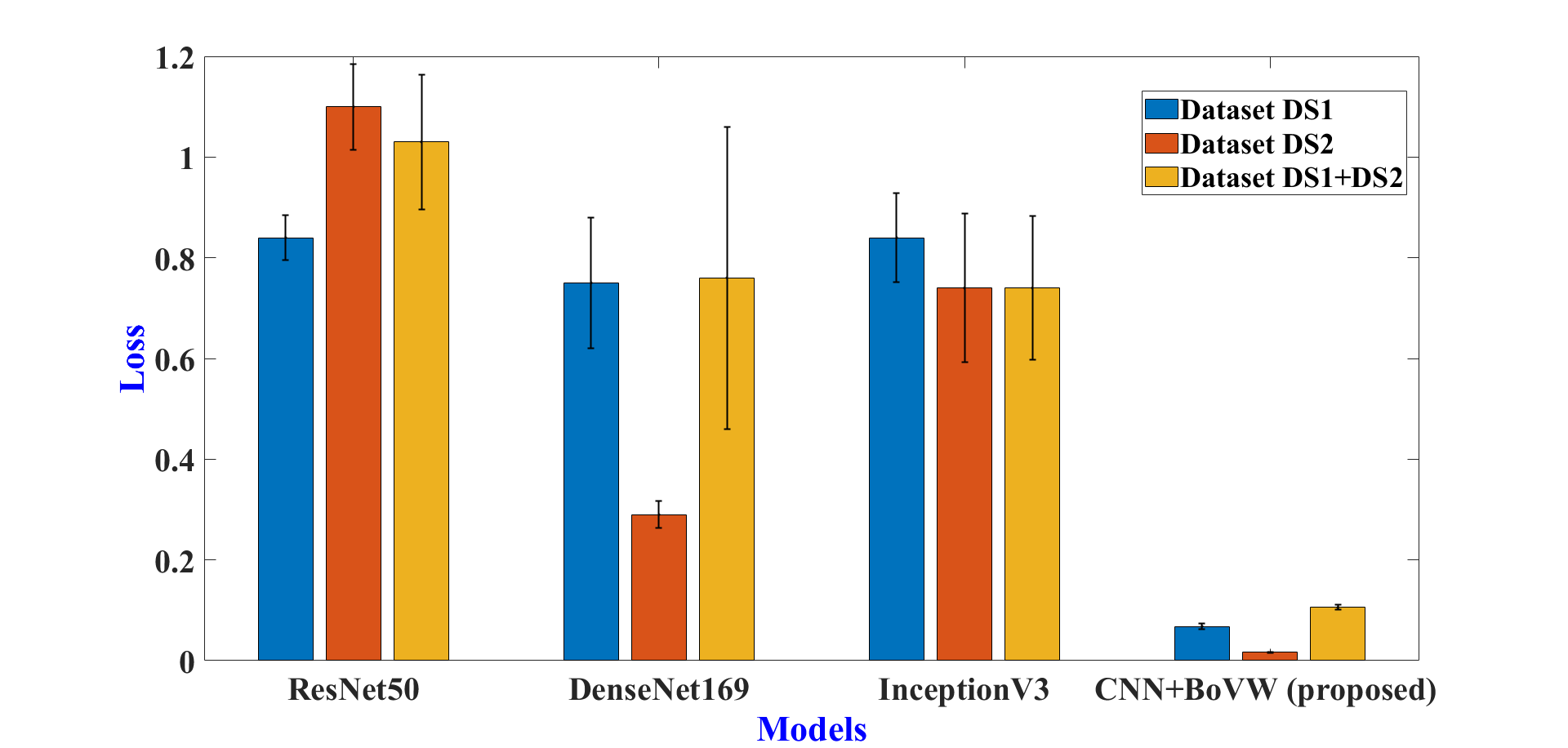}
\caption{Comparative  patch wise classification Cross Entropy Loss obtained with respect to each dataset. DS1- Microscopy dataset, DS2- WSI dataset. The losses are plotted as mean loss on y-axis with error bars showing standard error obtained after five fold cross validation. The losses are obtained after Experiments 1 and 2 as explained in the text.}
\label{subfig:fig2}
\end{figure*}

The graph in Fig. \ref{subfig:fig1} plots accuracy of each model tested on Dataset \textbf{DS1}, \textbf{DS2}, and merged Dataset \textbf{DS1 + DS2}. Accuracy of the proposed model is similar to DenseNet169 with respect to Dataset \textbf{DS1} but much higher in case of other two datasets.  Similarly,  Fig. \ref{subfig:fig2} plots the corresponding cross-entropy loss values. The loss for our proposed model is significantly low as compared to comparative models (ResNet50, DenseNet169, and InceptionV3). Also, it can be observed that there is no uniformity in the trend with respect to datasets.
\begin{table}[htbp]
\caption{Comparative Performance parameters for patch wise classification for combined Dataset \textbf{DS1 + DS2}. DS1- Microscopy dataset, DS2- WSI dataset}
\label{table2}
\centering
\renewcommand{\arraystretch}{2.5}
\begin{adjustbox}{width=\textwidth}
\begin{tabular}{|c|c|c|c|c|c|}
\hline
\textbf{Method}&\textbf{Precision(\%)}&\textbf{Recall(\%)}&\textbf{F1-Score(\%)}&\textbf{\shortstack{Multiclass \\ AUC(\%)}} \\
\hline
\multirow{1}{*}{\textbf{ResNet50 \cite{he2016deep}}}&69.95 $\pm$ 11.38&69.95 $\pm$ 11.38&69.95 $\pm$ 11.38&89.24 $\pm$ 6.66\\
\hline
\multirow{1}{*}{\textbf{InceptionV3 \cite{szegedy2015rethinking}}}&83.10 $\pm$ 9.30&83.10 $\pm$ 9.30&83.10 $\pm$ 9.30&95.39 $\pm$ 3.86\\
\hline
\multirow{1}{*}{\textbf{DenseNet169 \cite{huang2017densely}}}&89.26 $\pm$ 5.03&89.26 $\pm$ 5.03&89.26 $\pm$ 5.03&97.11 $\pm$ 3.05\\
\hline
\multirow{1}{*}{\textbf{\shortstack{CNN-BoVW \\ (Proposed)}}}&80.73 $\pm$ 2.84&80.32 $\pm$ 3.01&80.49 $\pm$ 2.96&95.67 $\pm$ 1.14\\
\hline
\end{tabular}
\end{adjustbox}
\end{table}

\par \textit{\textbf{Experiment 3:}} We know that there are different pyramid levels in WSI. So, we extracted the regions at three different pyramid levels as described in Section \ref{sec4.2}. These three levels were then passed through the model to test the sensitivity of the model towards lower resolution level patches when the details in the patch features are less sharp and indistinct to the machine. Table \ref{table3} shows the metrics obtained with respect to all three resolution levels. All the experiments are done on 5-fold cross-validation training and testing feature sets. From the Table \ref{table3}, after analyzing performance parameters at different levels of resolution, we deduce that our proposed model was less sensitive for scale 1 and the performance further declined for scale 2. This trend in performance signifies the importance of the presence of sharp details in biomedical images for better feature representation. Having high resolution details help to distinguish between different classes, especially as in our case where specific categorization of tumours between more than two classes is required.   
\begin{table*}[!htp]
\renewcommand{\arraystretch}{1.5}
\caption{Patch wise classification performance of CNN-BoVW model at different resolution levels (Experiment 3).}
\label{table3}
\centering
\begin{adjustbox}{width=\textwidth}
\begin{tabular}{|c|c|c|c|c|c|c|}
\hline
\textbf{Resolution}&\textbf{Precision}&\textbf{Recall}&\textbf{F1-Score}&\textbf{Multiclass AUC}&\textbf{Accuracy}&\textbf{Loss} \\
\hline
\multirow{1}{*}{\textbf{Scale 0}}&98.08\% $\pm$ 0.28\%&97.86\% $\pm$ 0.27\%&97.96\% $\pm$ 0.26\%&99.81\% $\pm$ 0.06\%&98.41\% $\pm$ 0.23\%&0.0165 $\pm$ 0.0008\\
\hline
\multirow{1}{*}{\textbf{Scale 1}}&91.59\% $\pm$ 1.83\%&89.25\% $\pm$ 1.88\%&90.20\% $\pm$ 1.75\%&99.15\% $\pm$ 0.69\%&97.48\% $\pm$ 0.47\%&0.0193 $\pm$ 0.0031\\
\hline
\multirow{1}{*}{\textbf{Scale 2}}&74.99\% $\pm$ 3.45\%&71.27\% $\pm$ 3.53\%&72.10\% $\pm$ 2.97\%&91.93\% $\pm$ 2.00\%&89.92\% $\pm$ 1.20\%&0.0863 $\pm$ 0.143\\
\hline
\end{tabular}
\end{adjustbox}
\end{table*} 

\label{sec4.4}
%
%
\section{Discussions}
\label{sec5}
Our proposed method has two main advantages; first, no heavy pre-processing or training deep model is required; second, the end-to-end model encapsulates the discriminative capability within the architectural framework for classification. The method was proposed keeping in mind some of the problems associated with patch level classification in respect to each class. For instance,  most of the area in the center of benign tumours are covered with white regions which, when divided into patches, can look like a part of both normal and benign class. Similarly, Invasive carcinoma is spread across the whole tissue and drawing a boundary around the tumour sometimes include all other classes, like normal regions containing soft tissue, stroma, and blood vessels, a small benign and \textit{In situ} tumour surrounded by malignant nuclei of invasive tumours. Therefore, when the patches are extracted from the bounding box around the boundary annotated by the expert, the traces of all other class types may remain with the invasive nuclei structures. In such cases, it becomes essential for the machine learning model to be able to distinguish among most related features from outliers (other class tumour patches) present in the set of patches extracted from the same region.  To tackle this problem of selecting features having low variance and discarding features with high variance, we developed a pipeline to extract deep learning features (of GoogleNet) followed by feature selection using Bag of Visual Words algorithm. We did not fine-tune the GoogleNet model for extracting features on our dataset and broadly extracted features from the pre-trained weights of ImageNet dataset. The goal was to avoid time to fine-tune the deep model on our dataset. The Bag of Visual Words model hence played the central role. \\
From Table 1, we see that all the state of the art deep learning architectures have performed moderately on the two datasets. Their metrics, when compared with deeper methods, do not show significant improvement. Moreover, their architecture does not provide flexibility to include discriminative capability within the model layers. \\
BoVW as a feature selection method created the marked difference between the two methods. The results reflect the fact that deep learning alone, even after fine-tuning, cannot be regarded as an optimal method for the classification of complex data \citep{tripathi2018histopathological}. To train a deep learning model from scratch in our case was not an option due to lack of enough sample points for the model to be able to generalize well. Further, with our model, using deep networks pre-trained on ImageNet to extract features was a design choice to be able to extract generalized features such as nuclei edges and boundaries, the density of stroma over nuclei structures, color and intensity information, and texture. Generally, nuclei structures present in the images are more sensitive to filter convolutions and yield high filter responses against background stromal structures and soft tissues because of their high color intensity, chromatin texture, and density. Therefore, fine-tuning for transfer learning was avoided to reduce cost and time of training heavy GoogleNet architecture.  The extracted features were then strategically selected and clustered per region, instead of extracting globally from the whole dataset so that the features could form a correlation among region patches. This way, through results, we demonstrated the enhancement in the diagnostic performance of the model for classifying breast tumours and normal regions in both microscopy and Whole Slide Images. \\
Although, the proposed method show improved performance and robustness to multiple datasets, it could not be extended for a clinical integration as a complete system. The method's applicability is limited as only a module for a larger system that classifies the disease in a larger context and not just at patch level. The robustness of the proposed model required for diagnosing tumours in clinical applications may be improved using a larger dataset containing varied patient biopsies, heterogeneous acquisition, staining, and scanning conditions of obtaining digitized dataset and more context-driven model that take into account the history of patch features from the same region to enable better feature correlation. 
\par For this work we used NVIDIA GeForce GTX 1080 Ti GPU and TitanX (PASCAL) GPU with 11GB RAM each for training our state of the art deep models \cite{huang2017densely,szegedy2015rethinking,he2016deep}. For our proposed model, we did not require GPU and used HP 280 G3 MT with 16 GB RAM  having Intel(R) Core(TM) i7-7700 CPU @ 3.60GHz, 3601 Mhz, 4 Core(s), 8 Logical Processors.

\section{Conclusion}
This work was motivated by the idea of proposing a robust and hybrid model that utilizes the key features obtained after deep model feature extraction. The framework provided an easy and flexible patch-based classification system for classifying breast tumours into four classes.    The flexibility of the system arises from the fact that any pre-trained deep learning model could be used to extract features without the need for fine-tuning and hence avoiding time and space. Such systems are robust due to their advantage in being reliable systems even if the dataset sample size is less, which is a vastly understood problem in the medical image research community.  With simple architectural changes, and the idea of including tradition algorithms for feature engineering helped us achieve a classification accuracy of 93.88\% for Microscopy images (\textbf{DS1}) and  98.41\% for WSI (\textbf{DS2}). The method could be further improved for image-wise classification for both WSI and Microscopy images.  
 

\section*{Acknowledgments}
This research was carried out in Indian Institute of Information Technology, Allahabad and supported, in part, by the Ministry of Human Resource and Development, Government of India and the Biomedical Research Council of the Agency for Science, Technology, and Research, Singapore. We are also grateful to the NVIDIA corporation for supporting our research in this area by granting us TitanX (PASCAL) GPU.

\bibliographystyle{unsrt}
\bibliography{classificationPatches}

\begin{thebibliography}{10}

\bibitem{waldum2008classification}
Helge~L Waldum, Arne~K Sandvik, Eiliv Brenna, Reidar Fossmark, Gunnar Qvigstad,
  and Jun Soga.
\newblock Classification of tumours.
\newblock {\em Journal of Experimental \& Clinical Cancer Research}, 27(1):70,
  2008.

\bibitem{apple2016sentinel}
Sophia~K Apple.
\newblock Sentinel lymph node in breast cancer: review article from a
  pathologist’s point of view.
\newblock {\em Journal of pathology and translational medicine}, 50(2):83,
  2016.

\bibitem{connolly2006changes}
James~L Connolly.
\newblock Changes and problematic areas in interpretation of the ajcc cancer
  staging manual, for breast cancer.
\newblock {\em Archives of pathology \& laboratory medicine}, 130(3):287--291,
  2006.

\bibitem{cserni2008variations}
G{\'a}bor Cserni, Simonetta Bianchi, Vania Vezzosi, Paul van Diest, Carolien
  van Deurzen, Istv{\'a}n Sejben, Peter Regitnig, Martin Asslaber, Maria~P
  Foschini, Anna Sapino, et~al.
\newblock Variations in sentinel node isolated tumour cells/micrometastasis and
  non-sentinel node involvement rates according to different interpretations of
  the tnm definitions.
\newblock {\em European Journal of Cancer}, 44(15):2185--2191, 2008.

\bibitem{vestjens2012relevant}
JHMJ Vestjens, MJ~Pepels, Maaike de~Boer, George~F Borm, Carolien~HM van
  Deurzen, Paul~J van Diest, JAAM Van~Dijck, EMM Adang, Johan~WR Nortier, EJ~Th
  Rutgers, et~al.
\newblock Relevant impact of central pathology review on nodal classification
  in individual breast cancer patients.
\newblock {\em Annals of oncology}, 23(10):2561--2566, 2012.

\bibitem{spanhol2016breast}
Fabio~Alexandre Spanhol, Luiz~S Oliveira, Caroline Petitjean, and Laurent
  Heutte.
\newblock Breast cancer histopathological image classification using
  convolutional neural networks.
\newblock In {\em 2016 international joint conference on neural networks
  (IJCNN)}, pages 2560--2567. IEEE, 2016.

\bibitem{gecer2018detection}
Baris Gecer, Selim Aksoy, Ezgi Mercan, Linda~G Shapiro, Donald~L Weaver, and
  Joann~G Elmore.
\newblock Detection and classification of cancer in whole slide breast
  histopathology images using deep convolutional networks.
\newblock {\em Pattern recognition}, 84:345--356, 2018.

\bibitem{huang2011time}
Chao-Hui Huang, Antoine Veillard, Ludovic Roux, Nicolas Lom{\'e}nie, and Daniel
  Racoceanu.
\newblock Time-efficient sparse analysis of histopathological whole slide
  images.
\newblock {\em Computerized medical imaging and graphics}, 35(7-8):579--591,
  2011.

\bibitem{doyle2006boosting}
Scott Doyle, Anant Madabhushi, Michael Feldman, and John Tomaszeweski.
\newblock A boosting cascade for automated detection of prostate cancer from
  digitized histology.
\newblock In {\em International conference on medical image computing and
  computer-assisted intervention}, pages 504--511. Springer, 2006.

\bibitem{basavanhally2013multi}
Ajay Basavanhally, Shridar Ganesan, Michael Feldman, Natalie Shih, Carolyn
  Mies, John Tomaszewski, and Anant Madabhushi.
\newblock Multi-field-of-view framework for distinguishing tumor grade in er+
  breast cancer from entire histopathology slides.
\newblock {\em IEEE transactions on biomedical engineering}, 60(8):2089--2099,
  2013.

\bibitem{bejnordi2016automated}
Babak~Ehteshami Bejnordi, Maschenka Balkenhol, Geert Litjens, Roland Holland,
  Peter Bult, Nico Karssemeijer, and Jeroen~AWM Van Der~Laak.
\newblock Automated detection of dcis in whole-slide h\&e stained breast
  histopathology images.
\newblock {\em IEEE transactions on medical imaging}, 35(9):2141--2150, 2016.

\bibitem{bahlmann2012automated}
Claus Bahlmann, Amar Patel, Jeffrey Johnson, Jie Ni, Andrei Chekkoury,
  Parmeshwar Khurd, Ali Kamen, Leo Grady, Elizabeth Krupinski, Anna Graham,
  et~al.
\newblock Automated detection of diagnostically relevant regions in h\&e
  stained digital pathology slides.
\newblock In {\em Medical Imaging 2012: Computer-Aided Diagnosis}, volume 8315,
  page 831504. International Society for Optics and Photonics, 2012.

\bibitem{wang2018breast}
Yaqi Wang, Lingling Sun, Kaiqiang Ma, and Jiannan Fang.
\newblock Breast cancer microscope image classification based on cnn with image
  deformation.
\newblock In {\em International Conference Image Analysis and Recognition},
  pages 845--852. Springer, 2018.

\bibitem{bandi2018detection}
Peter Bandi, Oscar Geessink, Quirine Manson, Marcory Van~Dijk, Maschenka
  Balkenhol, Meyke Hermsen, Babak~Ehteshami Bejnordi, Byungjae Lee, Kyunghyun
  Paeng, Aoxiao Zhong, et~al.
\newblock From detection of individual metastases to classification of lymph
  node status at the patient level: the camelyon17 challenge.
\newblock {\em IEEE transactions on medical imaging}, 38(2):550--560, 2018.

\bibitem{kausar2019hwdcnn}
Tasleem Kausar, MingJiang Wang, Muhammad Idrees, and Yun Lu.
\newblock Hwdcnn: Multi-class recognition in breast histopathology with haar
  wavelet decomposed image based convolution neural network.
\newblock {\em Biocybernetics and Biomedical Engineering}, 39(4):967--982,
  2019.

\bibitem{jonnalagedda2018regular}
Padmaja Jonnalagedda, Daniel Schmolze, and Bir Bhanu.
\newblock [regular paper] mvpnets: Multi-viewing path deep learning neural
  networks for magnification invariant diagnosis in breast cancer.
\newblock In {\em 2018 IEEE 18th International Conference on Bioinformatics and
  Bioengineering (BIBE)}, pages 189--194. IEEE, 2018.

\bibitem{mahbod2018breast}
Amirreza Mahbod, Isabella Ellinger, Rupert Ecker, {\"O}rjan Smedby, and
  Chunliang Wang.
\newblock Breast cancer histological image classification using fine-tuned deep
  network fusion.
\newblock In {\em International Conference Image Analysis and Recognition},
  pages 754--762. Springer, 2018.

\bibitem{graham2018classification}
Simon Graham, Muhammad Shaban, Talha Qaiser, Navid~Alemi Koohbanani, Syed~Ali
  Khurram, and Nasir Rajpoot.
\newblock Classification of lung cancer histology images using patch-level
  summary statistics.
\newblock In {\em Medical Imaging 2018: Digital Pathology}, volume 10581, page
  1058119. International Society for Optics and Photonics, 2018.

\bibitem{imagenet_cvpr09}
J.~Deng, W.~Dong, R.~Socher, L.-J. Li, K.~Li, and L.~Fei-Fei.
\newblock {ImageNet: A Large-Scale Hierarchical Image Database}.
\newblock In {\em CVPR09}, 2009.

\bibitem{tripathi2018histopathological}
Suvidha Tripathi and Satish Singh.
\newblock Histopathological image classification: Defying deep architectures on
  complex data.
\newblock In {\em International Conference on Recent Trends in Image Processing
  and Pattern Recognition}, pages 361--370. Springer, 2018.

\bibitem{ozdemir2011resampling}
Erdem Ozdemir, Cenk Sokmensuer, and Cigdem Gunduz-Demir.
\newblock A resampling-based markovian model for automated colon cancer
  diagnosis.
\newblock {\em IEEE transactions on biomedical engineering}, 59(1):281--289,
  2011.

\bibitem{difranco2011ensemble}
Matthew~D DiFranco, Gillian O’Hurley, Elaine~W Kay, R~William~G Watson, and
  Padraig Cunningham.
\newblock Ensemble based system for whole-slide prostate cancer probability
  mapping using color texture features.
\newblock {\em Computerized medical imaging and graphics}, 35(7-8):629--645,
  2011.

\bibitem{monaco2010high}
James~P Monaco, John~E Tomaszewski, Michael~D Feldman, Ian Hagemann, Mehdi
  Moradi, Parvin Mousavi, Alexander Boag, Chris Davidson, Purang Abolmaesumi,
  and Anant Madabhushi.
\newblock High-throughput detection of prostate cancer in histological sections
  using probabilistic pairwise markov models.
\newblock {\em Medical image analysis}, 14(4):617--629, 2010.

\bibitem{krishnan2012hybrid}
M~Muthu~Rama Krishnan, Chandan Chakraborty, Ranjan~Rashmi Paul, and Ajoy~K Ray.
\newblock Hybrid segmentation, characterization and classification of basal
  cell nuclei from histopathological images of normal oral mucosa and oral
  submucous fibrosis.
\newblock {\em Expert Systems with Applications}, 39(1):1062--1077, 2012.

\bibitem{doyle2008automated}
Scott Doyle, Shannon Agner, Anant Madabhushi, Michael Feldman, and John
  Tomaszewski.
\newblock Automated grading of breast cancer histopathology using spectral
  clustering with textural and architectural image features.
\newblock In {\em 2008 5th IEEE International Symposium on Biomedical Imaging:
  From Nano to Macro}, pages 496--499. IEEE, 2008.

\bibitem{krishnan2012automated}
M~Muthu~Rama Krishnan, Vikram Venkatraghavan, U~Rajendra Acharya, Mousumi Pal,
  Ranjan~Rashmi Paul, Lim~Choo Min, Ajoy~Kumar Ray, Jyotirmoy Chatterjee, and
  Chandan Chakraborty.
\newblock Automated oral cancer identification using histopathological images:
  a hybrid feature extraction paradigm.
\newblock {\em Micron}, 43(2-3):352--364, 2012.

\bibitem{wang2015exploring}
Daihou Wang, David~J Foran, Jian Ren, Hua Zhong, Isaac~Y Kim, and Xin Qi.
\newblock Exploring automatic prostate histopathology image gleason grading via
  local structure modeling.
\newblock In {\em 2015 37th Annual International Conference of the IEEE
  Engineering in Medicine and Biology Society (EMBC)}, pages 2649--2652. IEEE,
  2015.

\bibitem{caicedo2009histopathology}
Juan~C Caicedo, Angel Cruz, and Fabio~A Gonzalez.
\newblock Histopathology image classification using bag of features and kernel
  functions.
\newblock In {\em Conference on Artificial Intelligence in Medicine in Europe},
  pages 126--135. Springer, 2009.

\bibitem{kather2016multi}
Jakob~Nikolas Kather, Cleo-Aron Weis, Francesco Bianconi, Susanne~M Melchers,
  Lothar~R Schad, Timo Gaiser, Alexander Marx, and Frank~Gerrit Z{\"o}llner.
\newblock Multi-class texture analysis in colorectal cancer histology.
\newblock {\em Scientific reports}, 6:27988, 2016.

\bibitem{hou2016patch}
Le~Hou, Dimitris Samaras, Tahsin~M Kurc, Yi~Gao, James~E Davis, and Joel~H
  Saltz.
\newblock Patch-based convolutional neural network for whole slide tissue image
  classification.
\newblock In {\em Proceedings of the IEEE Conference on Computer Vision and
  Pattern Recognition}, pages 2424--2433, 2016.

\bibitem{fondon2018automatic}
Irene Fond{\'o}n, Auxiliadora Sarmiento, Ana~Isabel Garc{\'\i}a, Mar{\'\i}a
  Silvestre, Catarina Eloy, Ant{\'o}nio Pol{\'o}nia, and Paulo Aguiar.
\newblock Automatic classification of tissue malignancy for breast carcinoma
  diagnosis.
\newblock {\em Computers in biology and medicine}, 96:41--51, 2018.

\bibitem{nanni2018ensemble}
Loris Nanni, Stefano Ghidoni, and Sheryl Brahnam.
\newblock Ensemble of convolutional neural networks for bioimage
  classification.
\newblock {\em Applied Computing and Informatics}, 2018.

\bibitem{lee2018robust}
Byungjae Lee and Kyunghyun Paeng.
\newblock A robust and effective approach towards accurate metastasis detection
  and pn-stage classification in breast cancer.
\newblock In {\em International Conference on Medical Image Computing and
  Computer-Assisted Intervention}, pages 841--850. Springer, 2018.

\bibitem{spanhol2015dataset}
Fabio~A Spanhol, Luiz~S Oliveira, Caroline Petitjean, and Laurent Heutte.
\newblock A dataset for breast cancer histopathological image classification.
\newblock {\em IEEE Transactions on Biomedical Engineering}, 63(7):1455--1462,
  2015.

\bibitem{veeling2018rotation}
Bastiaan~S Veeling, Jasper Linmans, Jim Winkens, Taco Cohen, and Max Welling.
\newblock Rotation equivariant cnns for digital pathology.
\newblock In {\em International Conference on Medical image computing and
  computer-assisted intervention}, pages 210--218. Springer, 2018.

\bibitem{bejnordi2017diagnostic}
Babak~Ehteshami Bejnordi, Mitko Veta, Paul~Johannes Van~Diest, Bram
  Van~Ginneken, Nico Karssemeijer, Geert Litjens, Jeroen~AWM Van Der~Laak,
  Meyke Hermsen, Quirine~F Manson, Maschenka Balkenhol, et~al.
\newblock Diagnostic assessment of deep learning algorithms for detection of
  lymph node metastases in women with breast cancer.
\newblock {\em Jama}, 318(22):2199--2210, 2017.

\bibitem{aresta2019bach}
Guilherme Aresta, Teresa Ara{\'u}jo, Scotty Kwok, Sai~Saketh Chennamsetty,
  Mohammed Safwan, Varghese Alex, Bahram Marami, Marcel Prastawa, Monica Chan,
  Michael Donovan, et~al.
\newblock Bach: Grand challenge on breast cancer histology images.
\newblock {\em Medical image analysis}, 2019.

\bibitem{bach2018}
ICIAR 2018.
\newblock Bach.
\newblock \url{https://iciar2018-challenge.grand-challenge.org/Home/}, November
  2018.

\bibitem{xie2019deep}
Juanying Xie, Ran Liu, IV~Luttrell, Chaoyang Zhang, et~al.
\newblock Deep learning based analysis of histopathological images of breast
  cancer.
\newblock {\em Frontiers in genetics}, 10:80, 2019.

\bibitem{szegedy2017inception}
Christian Szegedy, Sergey Ioffe, Vincent Vanhoucke, and Alexander~A Alemi.
\newblock Inception-v4, inception-resnet and the impact of residual connections
  on learning.
\newblock In {\em Thirty-First AAAI Conference on Artificial Intelligence},
  2017.

\bibitem{han2017breast}
Zhongyi Han, Benzheng Wei, Yuanjie Zheng, Yilong Yin, Kejian Li, and Shuo Li.
\newblock Breast cancer multi-classification from histopathological images with
  structured deep learning model.
\newblock {\em Scientific reports}, 7(1):4172, 2017.

\bibitem{araujo2017classification}
Teresa Ara{\'u}jo, Guilherme Aresta, Eduardo Castro, Jos{\'e} Rouco, Paulo
  Aguiar, Catarina Eloy, Ant{\'o}nio Pol{\'o}nia, and Aur{\'e}lio Campilho.
\newblock Classification of breast cancer histology images using convolutional
  neural networks.
\newblock {\em PloS one}, 12(6):e0177544, 2017.

\bibitem{sirinukunwattana2016locality}
Korsuk Sirinukunwattana, Shan e~Ahmed~Raza, Yee-Wah Tsang, David~RJ Snead,
  Ian~A Cree, and Nasir~M Rajpoot.
\newblock Locality sensitive deep learning for detection and classification of
  nuclei in routine colon cancer histology images.
\newblock {\em IEEE Trans. Med. Imaging}, 35(5):1196--1206, 2016.

\bibitem{huang2017densely}
Gao Huang, Zhuang Liu, Laurens Van Der~Maaten, and Kilian~Q Weinberger.
\newblock Densely connected convolutional networks.
\newblock In {\em Proceedings of the IEEE conference on computer vision and
  pattern recognition}, pages 4700--4708, 2017.

\bibitem{chollet2017xception}
Fran{\c{c}}ois Chollet.
\newblock Xception: Deep learning with depthwise separable convolutions.
\newblock In {\em Proceedings of the IEEE conference on computer vision and
  pattern recognition}, pages 1251--1258, 2017.

\bibitem{simonyan2014very}
Karen Simonyan and Andrew Zisserman.
\newblock Very deep convolutional networks for large-scale image recognition.
\newblock {\em arXiv preprint arXiv:1409.1556}, 2014.

\bibitem{zhang2018classification}
Jianpeng Zhang, Yong Xia, Yutong Xie, Michael Fulham, and David~Dagan Feng.
\newblock Classification of medical images in the biomedical literature by
  jointly using deep and handcrafted visual features.
\newblock {\em IEEE journal of biomedical and health informatics},
  22(5):1521--1530, 2018.

\bibitem{wang2014mitosis}
Haibo Wang, Angel~Cruz Roa, Ajay~N Basavanhally, Hannah~L Gilmore, Natalie
  Shih, Mike Feldman, John Tomaszewski, Fabio Gonzalez, and Anant Madabhushi.
\newblock Mitosis detection in breast cancer pathology images by combining
  handcrafted and convolutional neural network features.
\newblock {\em Journal of Medical Imaging}, 1(3):034003, 2014.

\bibitem{ACM}
Suvidha Tripathi and Satish~Kumar Singh.
\newblock Cell nuclei classification in histopathological images using hybrid
  olconvnet.
\newblock {\em ACM Transactions on Multimedia Computing, Communications, and
  Applications}, 1(1):1--22, 2019.

\bibitem{krizhevsky2012imagenet}
Alex Krizhevsky, Ilya Sutskever, and Geoffrey~E Hinton.
\newblock Imagenet classification with deep convolutional neural networks.
\newblock In {\em Advances in neural information processing systems}, pages
  1097--1105, 2012.

\bibitem{szegedy2015going}
Christian Szegedy, Wei Liu, Yangqing Jia, Pierre Sermanet, Scott Reed, Dragomir
  Anguelov, Dumitru Erhan, Vincent Vanhoucke, and Andrew Rabinovich.
\newblock Going deeper with convolutions.
\newblock In {\em Proceedings of the IEEE conference on computer vision and
  pattern recognition}, pages 1--9, 2015.

\bibitem{harris1954distributional}
Zellig~S Harris.
\newblock Distributional structure.
\newblock {\em Word}, 10(2-3):146--162, 1954.

\bibitem{forgy1965cluster}
Edward~W Forgy.
\newblock Cluster analysis of multivariate data: efficiency versus
  interpretability of classifications.
\newblock {\em biometrics}, 21:768--769, 1965.

\bibitem{hand2001simple}
David~J Hand and Robert~J Till.
\newblock A simple generalisation of the area under the roc curve for multiple
  class classification problems.
\newblock {\em Machine learning}, 45(2):171--186, 2001.

\bibitem{he2016deep}
Kaiming He, Xiangyu Zhang, Shaoqing Ren, and Jian Sun.
\newblock Deep residual learning for image recognition.
\newblock In {\em Proceedings of the IEEE conference on computer vision and
  pattern recognition}, pages 770--778, 2016.

\bibitem{szegedy2015rethinking}
Christian Szegedy, Vincent Vanhoucke, Sergey Ioffe, Jonathon Shlens, and
  Zbigniew Wojna.
\newblock Rethinking the inception architecture for computer vision. arxiv
  2015.
\newblock {\em arXiv preprint arXiv:1512.00567}, 1512, 2015.

\bibitem{roy2019patch}
Kaushiki Roy, Debapriya Banik, Debotosh Bhattacharjee, and Mita Nasipuri.
\newblock Patch-based system for classification of breast histology images
  using deep learning.
\newblock {\em Computerized Medical Imaging and Graphics}, 71:90--103, 2019.

\end{thebibliography}

\end{document}